\documentclass[final]{cvpr}
\usepackage{times}
\usepackage{datetime}
\usepackage{epsfig}
\usepackage{graphicx}
\usepackage{amsmath}
\usepackage{amssymb}
\usepackage{upgreek}
\usepackage{graphicx}
\usepackage{dblfloatfix}    

\usepackage{makecell} 
\usepackage{pifont} 

\usepackage{threeparttable}
\usepackage[style=ieee, backend=biber, defernumbers=true]{biblatex}
\usepackage{caption}
\usepackage{bm} 
\newcommand{\ipr}{I_\text{pr}}
\newcommand{\isf}{I_\text{sf}}
\newcommand{\id}{I_\text{d}}
\newcommand{\ion}{I_\text{on}}
\newcommand{\ioff}{I_\text{off}}
\newcommand{\irefr}{I_\text{refr}}

\newcommand{\thr}{\theta}
\newcommand{\sens}{\Sigma}

\newcommand{\thon}{\thr_\text{on}}
\newcommand{\thoff}{\thr_\text{off}}
\newcommand{\thonoff}{\thr_\text{on,off}}

\newcommand{\trefr}{\Delta_\text{refr}}

\newcommand{\prbw}{B_\text{pr}}
\newcommand{\snr}{R_\text{S-N}}

\newcommand{\rlow}{\text{R}_\text{L}}
\newcommand{\rhigh}{\text{R}_\text{H}}

\newcommand{\rsignal}{R_\text{S}}
\newcommand{\rnoise}{R_\text{N}}
\newcommand{\rinput}{R_\text{I}}
\newcommand{\rnoiselimit}{\text{R}_\text{NL}}

\newcommand{\bbdelta}{\Delta_\text{bb}}
\newcommand{\bbhys}{\text{H}}
\newcommand{\bbrefr}{\text{t}_\text{bb}}
\newcommand{\bbignore}{\text{t}_\text{ignore}}

\newcommand{\tweak}{T}
\newcommand{\tweakthr}{T_{\thr}}
\newcommand{\tweakrefr}{T_\text{refr}}
\newcommand{\tweakbw}{T_\text{BW}}

\usepackage{ragged2e}
\newcolumntype{L}[1]{>{\RaggedRight\hspace{0pt}}p{#1}}
\newcolumntype{R}[1]{>{\RaggedLeft\hspace{0pt}}p{#1}}

\usepackage{kotex}

\usepackage[breaklinks=true,colorlinks,bookmarks=false]{hyperref}

\usepackage{enumitem} 
\setlist{nosep} 

\usepackage{xcolor}
\usepackage[printwatermark]{xwatermark}
\newwatermark*[page=1,color=gray!60,fontfamily=cmr,fontseries=m,scale=0.4,xpos=0,ypos=120]{This paper has been accepted for publication at the \\ IEEE/CVF Conference on Computer Vision and Pattern Recognition Workshops (CVPRW), Virtual, 2021}

\def\cvprPaperID{9}

\def\confYear{CVPR 2021}

\title{Feedback control of event cameras}

\author{Tobi Delbruck, Rui Graca, Marcin Paluch\\
Inst. of Neuroinformatics, UZH-ETH Zurich, Switzerland\\
{\tt\small Corresponding author: tobi@ini.uzh.ch}
}

\begin{document}

\twocolumn[{%
\renewcommand\twocolumn[1][]{#1}%
\maketitle

 \begin{minipage}[c]{0.67\textwidth}
    \includegraphics[width=\textwidth]{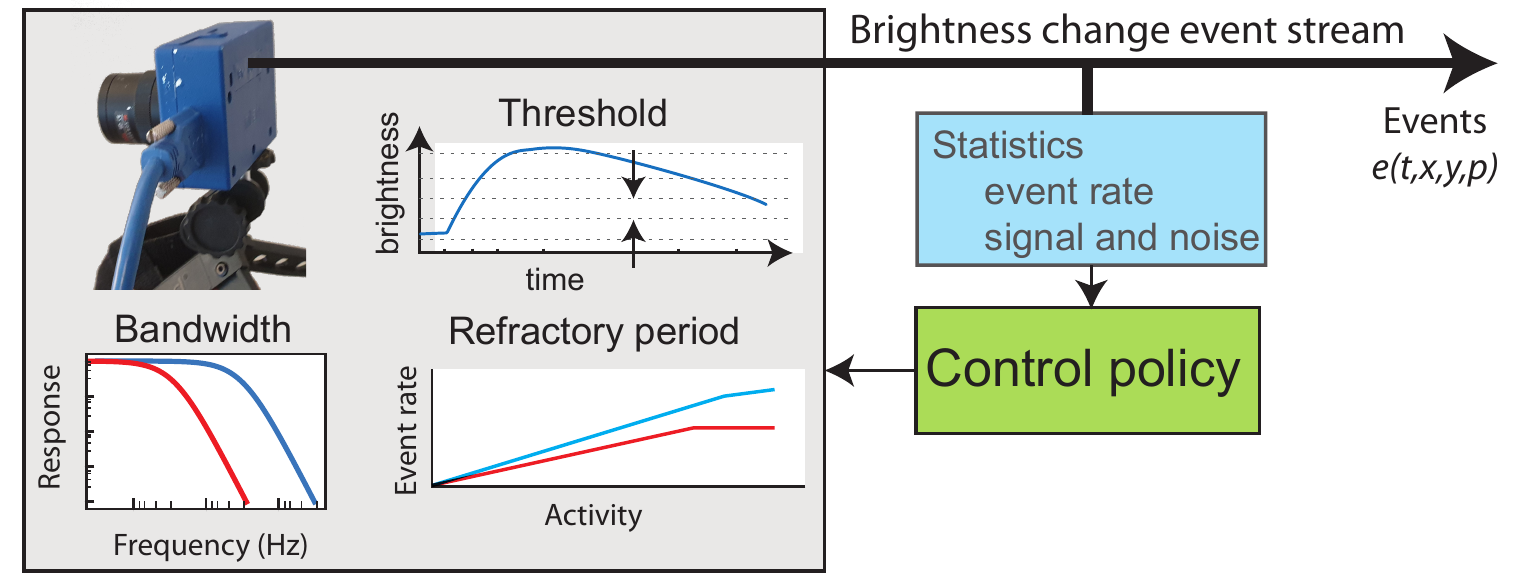}
  \end{minipage}\hfill
  \begin{minipage}[c]{0.3\textwidth}
    \captionof{figure}{
       Event camera feedback control dynamically modifies the
  threshold, bandwidth, and refractory period during operation to regulate the DVS output.
    } \label{fig:fig1}
  \end{minipage}
  
}]

\begin{abstract}
 Dynamic vision sensor event cameras produce a variable data rate stream of brightness change events. Event production
 at the pixel level is controlled by threshold, bandwidth, and refractory period bias current parameter settings. Biases must be adjusted to match application requirements and the optimal settings depend on many factors. As a first step towards automatic control of biases, this paper proposes fixed-step feedback controllers that use measurements of event rate and noise. The controllers regulate the event rate within an acceptable range using threshold and refractory period control, and regulate noise using bandwidth control. Experiments demonstrate model validity and feedback control.
\end{abstract}

\section{Introduction}

Dynamic vision sensor (\textbf{DVS}) event cameras~\cite{dvs128,survey-paper-dvs} 
such as those sold now by Inivation, Prophsee, Insightness,
and Celex offer superior dynamic range and sparse,
quick output that makes them useful for vision under 
uncontrolled lighting encountered in surveillance
and mobile applications. 
The high-quality output from standard cameras relies on decades of work to automatically optimize parameters such as 
column converter gain, noise filtering, exposure time, white balance, FPN correction, white/dark pixel removal, aperture control, and focus control~\cite{Nakamura2006-image-sensors-and-signal-proc-2006}. 
These algorithms provide high quality output from smartphone, surveillance, and automotive cameras.
There has been little work to dynamically optimize DVS output.
With the increasing use of DVS in uncontrolled environments
it is a good time to develop feedback control algorithms to optimize the DVS operating parameters. This study can guide the development of future DVS event cameras and their firmware and software frameworks. 

Fig.~\ref{fig:fig1} illustrates the control of threshold, refractory period, and bandwidth of a DVS camera based on statistical measurements of event rate and noise in the event stream.
The paper contributions and outline are:
\begin{enumerate}
    \item An overview of the targets for control (Sec.~\ref{sec:control_overview}).
    \item Models of DVS pixel bandwidth.
    threshold, and refractory period as functions of the pixel bias currents (Sec.~\ref{sec:dvsbiases}).
     \item Fixed-step regulators for DVS biases targeting event rate  and noise bounds (Sec.~\ref{sec:controller_design}), and an online estimate of signal and noise (Sec.~\ref{subsec:snrdef}).
    \item Experiments with a DVS camera that demonstrate model validity and control (Sec.~\ref{sec:results}).
\end{enumerate}
\raggedbottom
Symbols used throughout the paper are listed in Table~\ref{tab:symbols}.

\subsection{Prior work}
\label{subsec:priorwork}
Very high DVS event rates cause problems such as saturating the readout bus or hindering real time event processing. To overcome this, several schemes aiming to limit event rate were proposed.
Dropping DVS events that cannot be transmitted in time to the host computer by USB has been part of event sensor interface logic since~\cite{Berner2007-100dollar-aer-interface}. Ref.~\cite{Glover2018-controlled-delay-dvs} proposed an approach where the algorithm processing the packets used feedback control based on processing latency to adaptively adjust software algorithm processing costs, by adjusting the event batch size or fraction of dropped events.
A related approach \cite{Tapia2020-ASAP-ICRA} also proposed software to ensure that data can be processed in real time. They randomly discard a fraction of the events under feedback control to ensure real-time operation. In addition, they adaptively size the processed event packets to match the processing rate.

Recent industrial event cameras include on chip some form of control of the event readout or DVS frame sampling rate~\cite{Finateu2020-sony-prophesee,samsung-dvs-2020,Li-vlsi-symp-2019}.
By contrast, the DVS camera control described here controls the analog parameters of DVS pixel operation rather than the camera's event output bus or software processing algorithm. 
This way, it can control the event generation at the pixel level and optimize the pixel operation to a particular application or environment.

There appear to be no publications reporting performance of algorithms 
for automatically controlling the bias current parameters of DVS event cameras. 
Litzenberger~\cite{Litzenberger2007-car-counting} counted cars with a embedded camera using the first generation 64x64 pixel DVS chip~\cite{Lichtsteiner2005-iisw}. They used two sets of voltage biases optimized by hand for daytime and nighttime car counting where the thresholds were increased for nighttime (based on clock and calendar) to detect only car headlights.
A DVS threshold control algorithm was posted online around
2011\footnote{\href{https://sourceforge.net/p/jaer/code/HEAD/tree/jAER/trunk/src/ch/unizh/ini/jaer/chip/retina/DVS128BiasController.java}{DVS128BiasController.java on SourceForge.}}. Patent applications that appear to be based on this algorithm were filed in 2017 and 2018~\cite{Samsung-threshold-control-2017, Berner2018--bias-control}. A continuation in 2020~\cite{Berner2020-bias-control} extended one of the applications to vary the global sampling rate based on event rate.  However, none of these publications report experimental results about the effect of automatic bias control.

\begin{table}[t]
    \centering
    \begin{tabular}{rlp{4cm}}
        \textbf{Symbol} & \textbf{Default} & \textbf{Description}\\
        \hline
        \multicolumn{3}{l}{\textbf{Biases}--Sec.~\ref{sec:dvsbiases}}\\
        $\ipr$ & 1nA& Photoreceptor bias \\
        $\isf$ & 25pA&Source-follower bias \\
        $\id$ & 20nA & Differencing amp bias  \\
        $\ion$ &1.3uA & ON comparator bias \\
        $\ioff$ &300pA& OFF comparator bias  \\
        $\irefr$ &5nA & Refractory period bias  \\
        \hline
        \multicolumn{3}{l}{\textbf{Event threshold}--Sec.~\ref{subsec:threshold_theory}}\\
        $\thr$ & 0.28 e-fold/ev & Temporal contrast threshold \\
        $\sens$ & 3.6 ev/e-fold & Sensitivity\\
        \hline
        \multicolumn{3}{l}{\textbf{Bandwidth}--Sec.~\ref{subsec:bandwidth}}\\
        $\prbw$ &--& Photoreceptor bandwidth \\
        \hline
        \multicolumn{3}{l}{\textbf{Refractory period}--Sec.~\ref{subsec:refractoryperiod}}\\
         $\trefr$&$\approx$1us& Refractory period \\
        \hline
        \multicolumn{3}{l}{\textbf{Tweaks}--Sec.~\ref{sec:tweak}  with  $[1/\tweak_\text{min},\tweak_\text{max}]$ values}\\
        $\tweakthr$ & [1/10,10] & Threshold tweak \\
        $\tweakbw$ & [1/30,30] & Bandwidth tweak  \\
        $\tweakrefr$ & [1/100,8] & Refractory period tweak\\
        \hline
        \multicolumn{3}{l}{\textbf{Fixed-step control}--Sec.~\ref{sec:controller_design}}\\
        $\bbdelta$ & 0.1 & Tweak change. \\
        $\bbhys$ & 1.5 & Hysteresis; see Fig.~\ref{fig:fig3_event_rate_controllers}. \\        
        $\bbignore$ & 1s & Time to ignore events. \\
        $\bbrefr$ & 2s & Minimum control interval.\\
        \hline
        \multicolumn{3}{l}{\textbf{Event rate bounding and limiting}--Secs.~\ref{subsec:bangbangthresholdcontrol},\ref{subsec:refractory_period_control}}\\
        $\rhigh$ & 300kHz& Upper event rate limit\\
        $\rlow$  &100kHz & Lower event rate limit \\
        \hline
        \multicolumn{3}{l}{\textbf{Signal and Noise}--Sec.~\ref{subsec:bw_controller_design}}\\
        $\rinput$ &--& Input (raw) event rate\\
        $\rnoise$ &--& Noise event rate \\
        $\rsignal$ &--& Signal (cleaned) event rate\\
        $\rnoiselimit$ &0.5 Hz/pix& Limit for noise\\
         $\snr$ & --& Normalized S-N difference \\
         \hline
      \end{tabular}
    \caption{Symbols with values used in this paper.}
    \label{tab:symbols}
\end{table}

\begin{figure*}[hb]
  \centering
  \includegraphics[width=\textwidth]{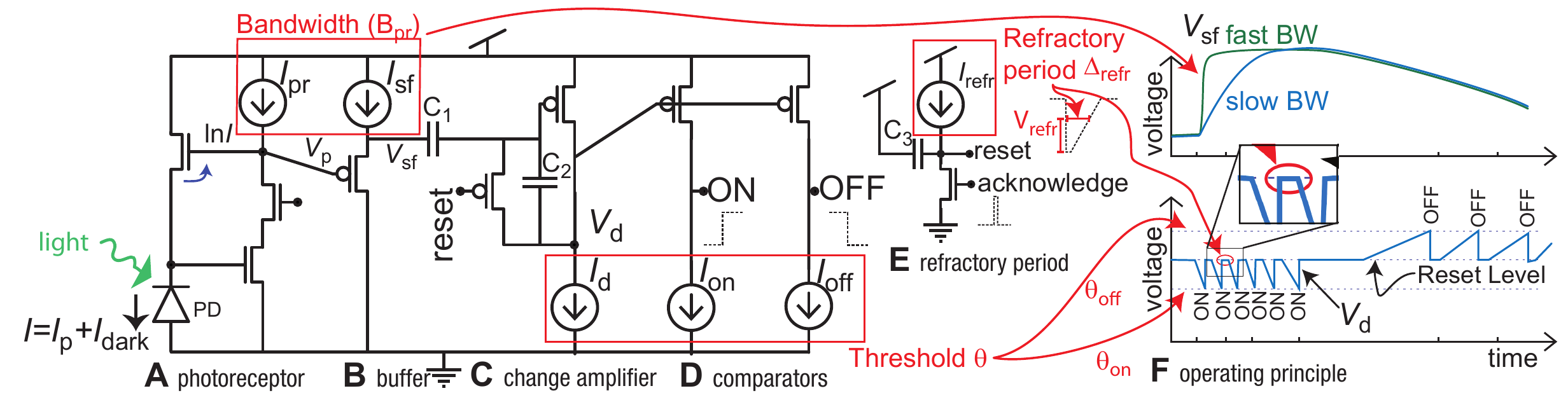}
  \captionof{figure}{DVS pixel circuit with biases. (Adapted from \cite{v2e}.)}
  \label{fig:fig2_pixelcircuit}
\end{figure*}

\section{Overview of control aims}
\label{sec:control_overview}

DVS event camera systems face at least two competing constraints: 
Adjusting biases to result in more events usually means more informative data,
but also more computing cost and more noise. We will assume there are two primary objectives of bias control:

\textbf{Objective 1 is to limit the event rate within acceptable bounds}.
We will assume that there is a largest acceptable rate of events that the DVS can output without bus saturation or that the system can process in real time. We call it the \textbf{high rate limit} $\rhigh$. 
Likewise, there is a \textbf{low rate limit} $\rlow$, which represents 
the quiescent event rate that the system can afford to process all the time. 
The job of the bias controller is to regulate the DVS event rate within these bounds. Sec.~\ref{sec:event_rate_control} explains our approach for regulating this bound.

\textbf{Objective 2 is to optimize the signal and noise.} It is important to realize it is not to maximize the signal to noise ratio, since this objective is easily achieved by removing almost all the noise by using a high threshold and small bandwidth---doing this also removes much of the informative signal events.  Sec.~\ref{subsec:snrdef} explains how we can measure the noise event rate $\rnoise$, and from it compute a signal versus noise quantity called $\snr$, Sec.~\ref{sec:limiting_noise_results} shows experimental data of $\snr$, and Sec.~\ref{sec:limiting_noise_results} demonstrates a controller that regulates $\rnoise$ to remain close to a rate $\rnoiselimit$.

\section{DVS biases}
\label{sec:dvsbiases}

Fig.~\ref{fig:fig2_pixelcircuit} shows a schematic of the DVS pixel with the associated bias current sources and sinks. The currents are specified at the pixel array level by the chip's bias generator. Changing the bias currents changes the 
\textbf{temporal contrast event threshold} $\bm{\thr}$,
the \textbf{photoreceptor bandwidth} $\bm{\prbw}$, 
and the \textbf{refractory period} $\bm{\trefr}$ between events. 

The critical threshold for creating ON and OFF events is controlled by the $\id$, $\ion$, and $\ioff$ currents.
The temporal bandwidth (speed) of the photoreceptor front end is controlled by $\ipr$ and $\isf$.
The refractory period between events---while the change amplifier is held in reset---is controlled by $\irefr$.
Together, these 5 bias currents control the most important characteristics of DVS pixels.
The larger the bias current, the higher the threshold or the faster the circuit.

\subsection{`Tweaking' biases around operating point}
\label{sec:tweak} 
The jAER developers realized that DVS users would need a user friendly control panel to control otherwise cryptic bias currents, and so developed the notion of a dimensionless bias `tweak' to control  DVS bias currents\footnote{\href{https://github.com/SensorsINI/jaer/blob/master/src/ch/unizh/ini/jaer/chip/retina/DVSTweaks.java}{DVSTweaks in jAER}}.  
The tweak interface - if properly optimised - automatically limits the range of bias modification, guarantees that the system stays within the functional range of its parameters, and hence assures safety of the controller. The DVS bias currents start from a nominal operating point (Table~\ref{tab:symbols}).
A linear tweak results in an exponential scaling of bias currents around the operating point.
The \textbf{threshold tweak} $\bm{\tweakthr}$, \textbf{refractory period tweak} $\bm{\tweakrefr}$, and \textbf{bandwidth tweak} $\bm{\tweakbw}$, map $\tweak$ to the actual bias current(s) range(s), such that the current $I$ is computed from
\begin{equation}
    I=I_0
    \begin{cases}
    \text{e}^{{\tweak \times \ln{\tweak_\text{max}}}} & \text{if $0\leq\tweak\leq +1$}\\
    \text{e}^{{\tweak \times \ln{\tweak_\text{min}}}} & \text{if $0>\tweak\geq -1$}\\
    \end{cases}
\end{equation}
where $I_0$ is the nominal current.
Limiting $\tweak$ to the range $\tweak\rightarrow[-1,1]$ limits the current to the functional operation range $I \rightarrow I_0\times [1/\tweak_\text{min},\tweak_\text{max}]$.
Table~\ref{tab:symbols} lists the $[1/\tweak_\text{min},\tweak_\text{max}]$ values used in this study, which were adjusted by the chip designers for functional DVS operation. 



\subsection{Temporal contrast threshold}
\label{subsec:threshold_theory}
Ref.~\cite{dvs-temperature-bias} showed that the event contrast thresholds $\thonoff$ 
which are the natural log intensity change 
for creating ON and OFF events are 
\begin{equation}
    \thonoff=A_\thr \ln{\frac{I_\text{on,off}}{\id}},
    \label{eq:threshold_from_currents}
\end{equation}
where for the DAVIS346 camera used in this study
\begin{equation}
A_\Theta\approx\frac{1.5 \text{C}_2}{ \text{C}_1}\approx 1/15.5.
\end{equation}
The $\thr$ magnitude can range over about
$[0.15,0.5]$.
For this range of small $\thr$, 
 $1+\thr$ is approximately the relative change of intensity to create an event. 
\eg, if $\thr=-0.2$,
then a decrease by a factor of about 0.8 creates an OFF event.

We will assume equal ON and OFF threshold, i.e.
$\thr=\thon=-\thoff$,
to balance the ON and OFF event rates. We define the \textbf{temporal contrast sensitivity} $\sens$:
\begin{equation}
    \sens=\frac{1}{\thr}.
\end{equation}
$\sens$  has units of events per e-fold intensity change.

We can use \eqref{eq:threshold_from_currents} to predict approximately
how changing $\thr$ will change the DVS event rate, 
if we can assume that the only thing that changes is the threshold.
Naively, we can expect that the event rate is simply scaled by $\sens$,
because the smaller the threshold,
the more events will be produced. However, if $\thr>\thr_\text{max}$, where $\thr_\text{max}$ is the highest contrast feature, then no events will be generated. If the scene contains a distribution of temporal contrasts, then increasing sensitivity will gradually expose more and more parts of the scene to create events. I.e., for low sensitivity, only the highest contrast features will create events.
Thus we expect that at least within some range, the total signal event rate
$R$ will follow the linear relationship
\begin{equation}
    R
    =\frac{R_0}{\sens_0-\sens_\text{min}}(\sens-\sens_\text{min}),
    \label{eq:ratevsthreshold}
\end{equation}
where $R_0$ 
is the nominal event rate at the nominal sensivity $\sens_0$ (with $\tweakthr=0$), and $\sens_\text{min}=1/\thr_\text{max}$ is the minimum sensitivity that triggers events. 
Secs.~\ref{subsec:ratevsthreshold} show that this model accurately describes experimental results for a useful range of $\sens$. For large $\sens$, measurements suggest that noise significantly contributes to the total event rate, making the linear model no longer valid.

\subsection{Photoreceptor bandwidth}
\label{subsec:bandwidth}
To first approximation,
the DVS pixel photoreceptor circuit can be considered as a 2nd-order lowpass filter, 
where the first stage cutoff is set by the photocurrent $I_\text{p}$ 
together with the photoreceptor bias current $\ipr$ 
and the 2nd stage cutoff is set by the source follower buffer bias current $\isf$.
For this paper,
we assume  
that the overall bandwidth $\prbw$ around some operating point
is a function $f(\ipr,\isf)$ of the two bias currents that increases monotonically with the currents:
\begin{equation}
    \prbw=B_{\text{pr}_{0}} f\left(\frac{\ipr}{I_{\text{pr}_{0}}},\frac{\isf}{I_{\text{sf}_{0}}}\right)
    \label{eq:bandwidth}
\end{equation}
where the 0 subscript means the nominal value. The function is complex and depends strongly on the absolute light intensity.

It is difficult to infer the effect of changing $\prbw$ on an arbitrary event stream.
We only know that increasing $\prbw$ will increase $R$ until the bandwidth exceeds all input signal frequencies, and will continue to increase $R$ by increasing noise events. Secs.~\ref{sec:snr_results} and \ref{sec:limiting_noise_results} show experimental results.

\subsection{Refractory period}
\label{subsec:refractoryperiod}
Fig.~\ref{fig:fig2_pixelcircuit} parts E\&F shows how $\irefr$ controls the dead time $\trefr$ between events. During this refractory period, the change amplifier is held in reset. The \textsl{reset} signal is pulled low by the pixel \textsl{acknowledge} signal, and while \textsl{reset} rises back up, the reset switch transistor $M_\text{r}$ shorts across the change amplifier output $V_\text{d}$ and input $V_\text{r}$, balancing it.
This refractory period $\trefr$ is determined by
\begin{equation}
    \trefr= \frac{\text{C}_\text{3}}{\irefr \text{V}_\text{refr}}
    \label{eq:refr_vs_current}
\end{equation}
where for the DAVIS346 used in this study, $C_\text{3}\approx 20\text{fF}$ and $V_\text{refr}\approx0.5\text{V}$; \eg $\trefr=10\text{ms}$ requires $\irefr=4\text{pA}$.

During the refractory period, 
the change amplifier ignores changes in the photoreceptor output. 
The refractory period sets a hard limit on the maximum event rate per pixel,
and it also reduces the effect of any input brightness change by discarding the changes that occur during the refractory period.

For a single pixel generating events at an instantaneous rate $r_0$, the time interval between events is $T_0=1/r_0$. A simple model assumes that the refractory period $\trefr$ effectively increases the time between events to $T_0+\trefr$, resulting in the new rate $r=1/({T_0+\trefr})$.
Averaged over the whole pixel array, the total rate $\rsignal$ is
\begin{equation}
    \rsignal = N_\text{total} \int_{0}^{\infty} \frac{f(T) dT}{T+\trefr}, 
    \label{eq:event_rate_refr}
\end{equation}
\noindent where $N_\text{total}$ is number of DVS pixels and $f(T)dT$ gives the fraction of pixels with time interval between events in the range $[T,T+dT]$. It follows that for any given $f(T)$ (i.e., scene with constant statistical character) the refractory period monotonically limits the event rate. Experimental confirmation and a control experiment are presented in Sections~\ref{sec:refr_effect_expt} and \ref{sec:refr_control_expt}.

\section{Controller design}
\label{sec:controller_design}
The previous section described simple models of event rate, signal and noise versus threshold, refractory period, and bandwidth biases. The main conclusion is that event rate and noise vary monotonically with bias. Therefore, all our controllers are very simple: They vary the relevant tweak by fixed steps until the objective is obtained. In this paper, we will call this \textit{fixed-step control}. Fixed-step control only requires monotonicity of the controller effect on the DVS output. Fixed-step control can take only small steps to result in accurate control, which makes it slow, but its easily obtained stable outcome may be an advantage for many applications.

\subsection{Controlling event rate}
\label{sec:event_rate_control}

\begin{figure}
    \centering
    \includegraphics[width=.8\columnwidth]{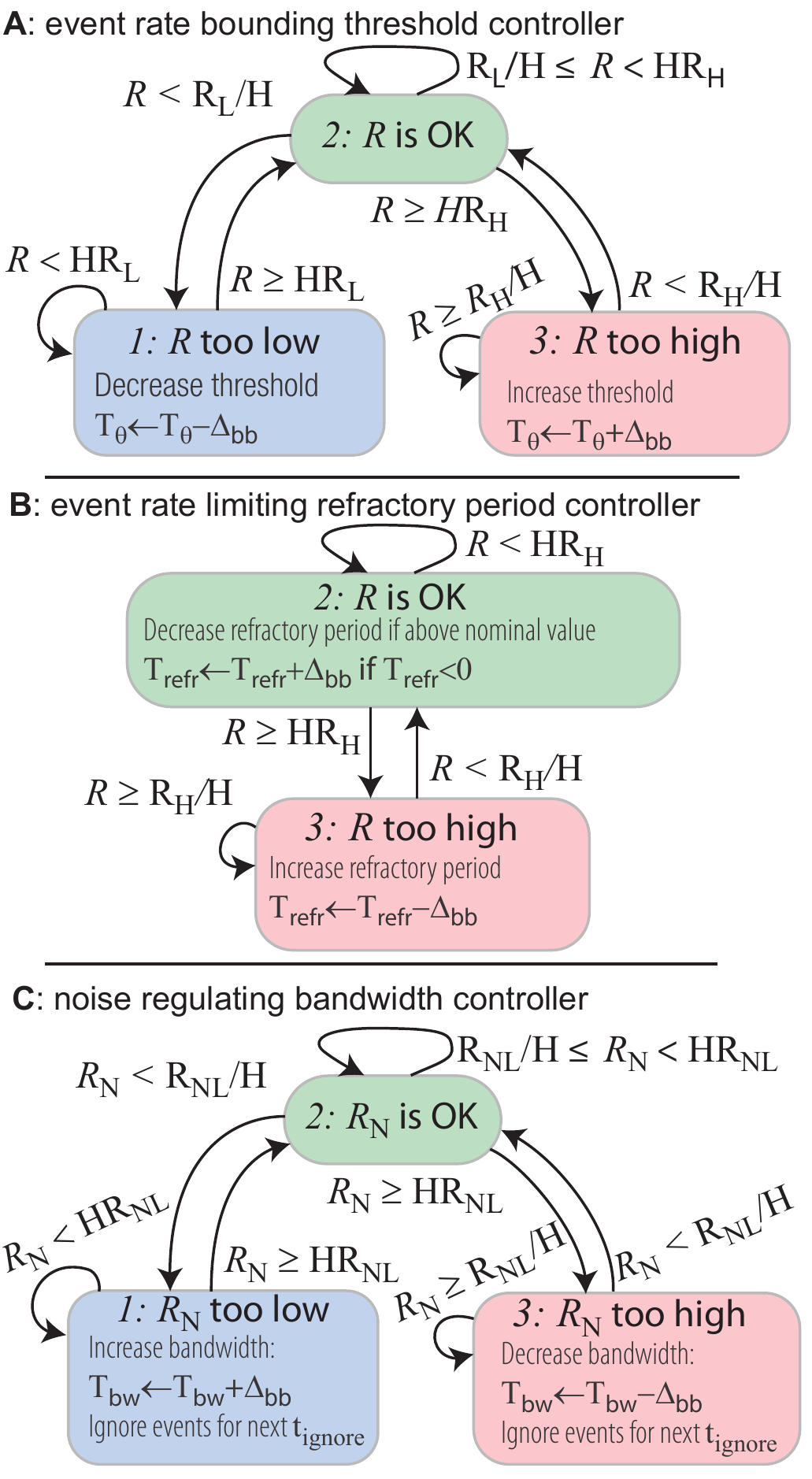}
    \caption{Event rate ($R$) and noise ($\rnoise$) fixed-step controllers.}
    \label{fig:fig3_event_rate_controllers}
\end{figure}

Fig~\ref{fig:fig3_event_rate_controllers} shows the design of the threshold and refractory period controllers that are described in the following. We measure event rate by a simple box filter that counts events during a period (usually 300\,ms) and divides the count by the period.

\subsubsection{Threshold control for bounding event rate}
\label{subsec:bangbangthresholdcontrol}
\vspace{-6pt}
The threshold controller shown in Fig.~\ref{fig:fig3_event_rate_controllers}A 
increases $\thr$ while the event rate $R$ is above the \textbf{upper bound} $\rhigh$, 
and decreases it while $R$ is below the \textbf{lower bound} $\rlow$. 
Every increase of $\sens$ creates noise, 
so the control occurs at a rate $1/\bbrefr$. 
A hysteresis factor $\bbhys$ holds the state until $R$ crosses the bound multiplied or divided by $\bbhys$.


\subsubsection{Refractory control for limiting event rate}
\label{subsec:refractory_period_control}
\vspace{-6pt}
The refractory period controller shown in Fig.~\ref{fig:fig3_event_rate_controllers}B 
limits the maximum event rate.
It increases $\trefr$ while the event rate $R$ is above the upper bound $\rhigh$, 
and gradually decreases it back to the default value once $R$ drops below $\rhigh$. 
In contrast to threshold control, changing refractory period does not produce much noise, but to reduce control actions, we use the same limits on control rate and hysteresis as for the threshold controller.

\subsection{Controlling signal and noise with bandwidth}
\label{subsec:bwcontrol}
\begin{figure}
    \centering
    \includegraphics[width=\columnwidth]{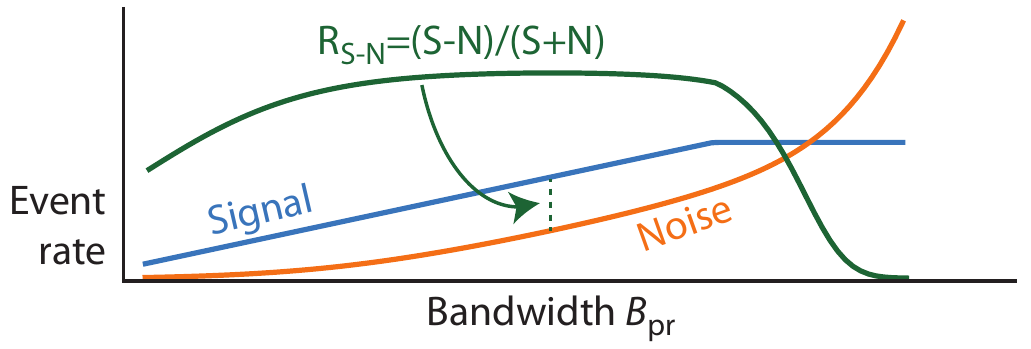}
    \caption{Concept for optimizing signal versus noise by controlling bandwidth. }
    \label{fig:bw_control_concept}
\end{figure}

Fig.~\ref{fig:bw_control_concept} illustrates the concept of bandwidth control for passing as much signal as possible without creating too much noise.  As we increase bandwidth, more and more of the high frequency signal components create brightness change events, but noise also increases. Eventually, the bandwidth is sufficient to capture all signal frequencies, while noise continues to increase. Sec.~\ref{sec:snr_results} shows an experiment for which an optimum setting exists, resulting in a maximum difference between signal and noise.

\subsubsection{Estimating signal and noise}
\label{subsec:snrdef}
\vspace{-6pt}
 We assume that signal and noise characteristics remain stable to allow their measurement.
Furthermore, we assume that a denoising algorithm can correctly filter out the DVS background activity noise, which 
lets us estimate a  measure of the \textbf{normalized signal minus noise difference} $\snr$ from 
\begin{equation}
\label{eq:snr}
    \snr=\frac{\rsignal-\rnoise}{\rsignal+\rnoise},
\end{equation}
where the event rates $\rsignal$ and $\rnoise$ are the rate of signal and noise events. In this study, we measure $\rnoise$ by using a correlation-based denoising algorithm called the Background Activity Filter~\cite{baf-filter-delbruck-frame-free}. If $\rinput$ is the input event rate, and $\rnoise$ is the rate of events that are removed by denoising, then  $\rsignal=\rinput-\rnoise$ is the rate of denoised signal events.
$\snr$ ranges over $[-1:1]$ and takes on a maximum value when the difference between signal and noise rates is maximal.
We adjusted the denoising correlation time to preserve the signal and remove most of the noise even at the highest bandwidth setting.
Sec.~\ref{sec:snr_results} shows experimental results of estimated $\snr$.

\subsubsection{Controlling bandwidth to regulate noise}
\label{subsec:bw_controller_design}
\vspace{-6pt}
A bandwidth controller might adjust bandwidth to maximize $\snr$. However, the online measurement of $\snr$ requires very long averaging to remove the effect of fluctuating signal characteristics, so we did not attempt to develop a controller. 
Instead, we developed the state-machine fixed-step controller to regulate noise illustrated in Fig.~\ref{fig:fig3_event_rate_controllers}C. It operates similarly to the threshold and refractory period controllers, but its aim is to keep the noise event rate close to a target $\rnoiselimit$.
Sec.~\ref{sec:limiting_noise_results} shows control experiment results.


\section{Experimental results}
\label{sec:results}
This section reports our measurements and control experiment results. 
Results are based on source code \textsl{DVSBiasController}\footnote{\href{https://github.com/SensorsINI/jaer/blob/master/src/ch/unizh/ini/jaer/chip/retina/DVSBiasController.java}{DVSBiasController.java in jAER on github}} using a prototype DAVIS346 camera from iniVation that uses a DAVIS sensor IC from our group~\cite{davis346-Taverni2018}.

\subsection{Experimental setup}
\label{sec:setup}

\begin{figure}
    \centering
    \includegraphics[width=.7\columnwidth]{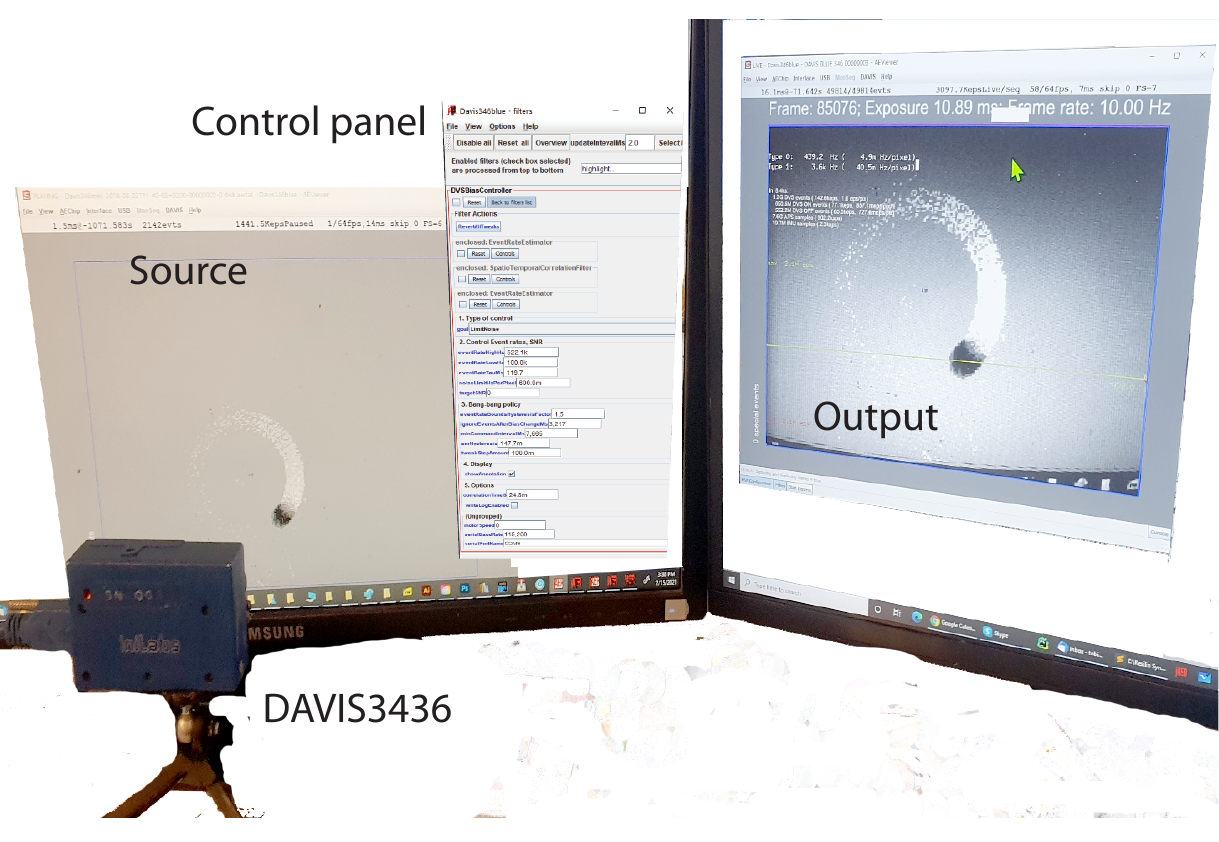}
    \caption{Setup for bias control experiments (except for Fig.~\ref{fig:event_rate_vs_refr}). The source monitor displays a stimulus whose activity rate and contrast can be easily controlled. The capture monitor displays the output of the Inivation DAVIS346 camera~\cite{davis346-Taverni2018} and the bias controller and tweaks interface. }
    \label{fig:setup}
\end{figure}

Fig.~\ref{fig:setup} shows the setup for experiments. The DAVIS346 camera
is connected to the host computer and the bias control runs on the host.
We control the DVS event rate by changing the speed and contrast of the displayed stimulus (here a recorded rotating dot).

\subsection{Effect of bias tweaks}
\label{sec:effect_of_bias_tweaks}
This section presents measurements of the effect of bias tweaks on event rate and signal versus noise.
\subsubsection{Effect of threshold on event rate}
\label{subsec:ratevsthreshold}

\begin{figure}
    \centering
    \includegraphics[width=\columnwidth]{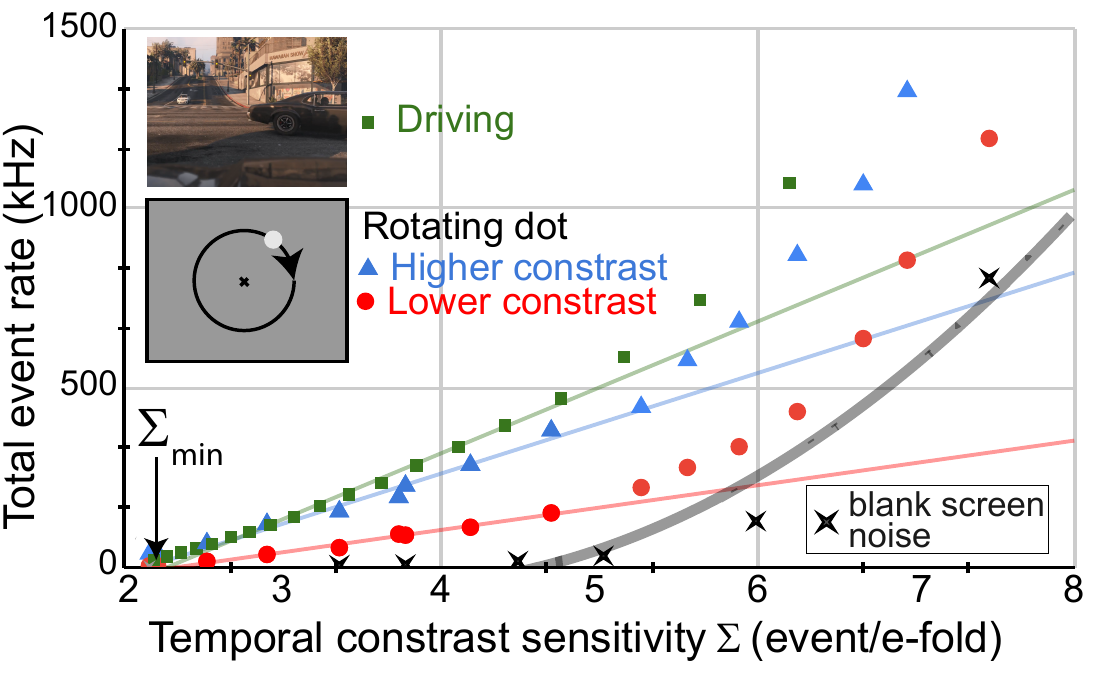}
    \caption{Measured total event rate versus estimated sensitivity ($\sens=1/\thr$) for rotating dot and driving scenes. }
    \label{fig:eventrate_vs_threshold}
\end{figure}

Estimating the event threshold from the bias currents allows plotting the measured event rate versus estimated threshold. Fig.~\ref{fig:eventrate_vs_threshold} compares the measured results with the theory of Sec.~\ref{subsec:threshold_theory} with a rotating dot and driving video.
The dot video is presented with two different contrasts.
 Over a sensitivity range of [2:5]\,event/e-fold there is a linear relationship; above this $\sens$ range, noise increases the event rate. The threshold sensitivity $\sens_\text{min}$ is about 2.2 for all of the tested inputs.
The gain factor depends on the visual input, but the intercept determined by the maximum contrast in the visual input, 
since contrast smaller than the threshold does not trigger events. 


\subsubsection{Effect of refractory period on limiting event rate}
\label{sec:refr_effect_expt}
\begin{figure}
    \centering
    \includegraphics[width=.8\columnwidth]{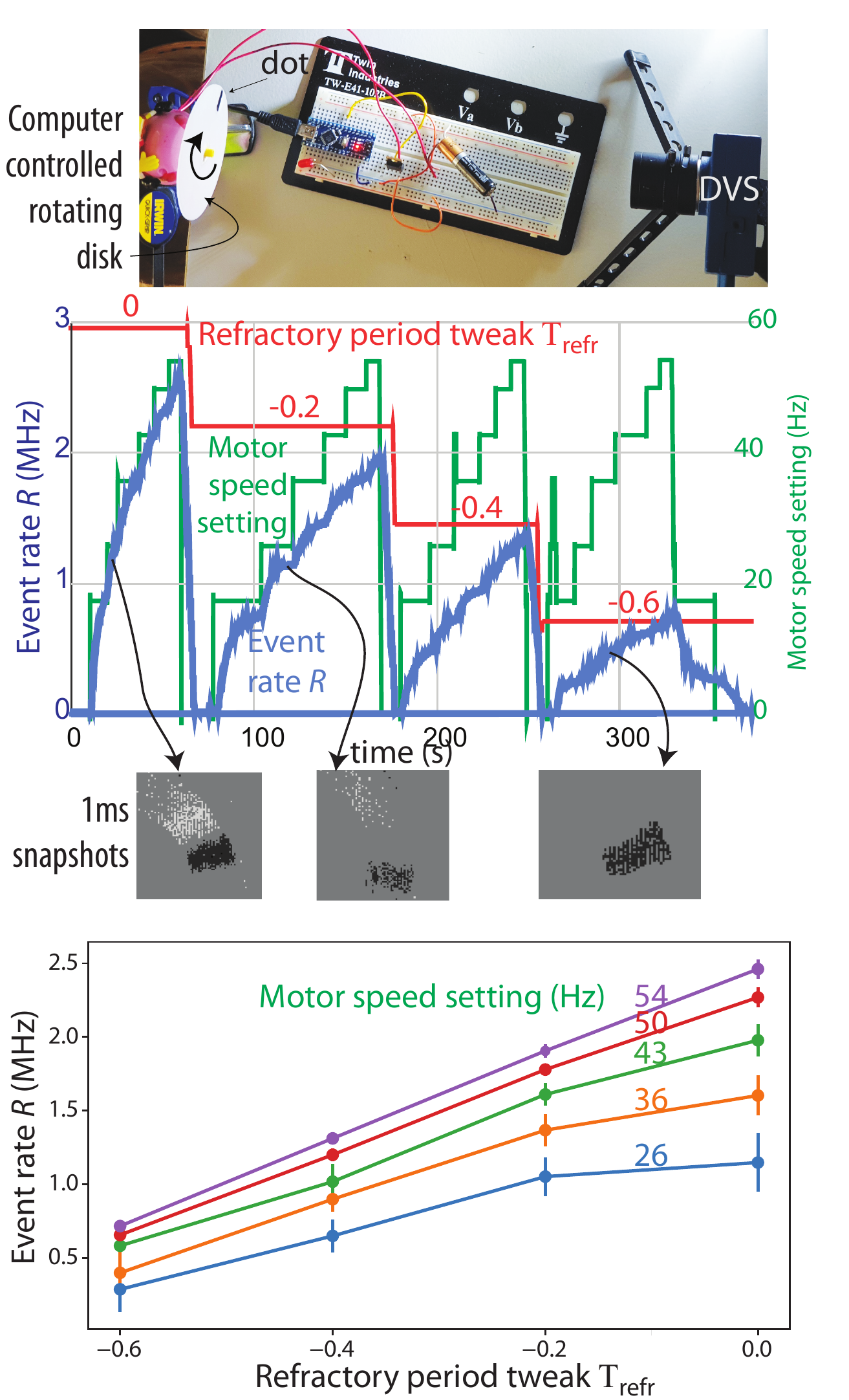}
    \caption{Measured event rate as a function of time and for different motor speed and refractory period tweak $\tweakrefr$ settings.  }
    \label{fig:event_rate_vs_refr}
\end{figure}

\vspace{-6pt}
Fig~\ref{fig:event_rate_vs_refr} shows limiting event rate with $\tweakrefr$ control. Since the refractory period affects the dead time after each event, we used a physical continuously rotating dot to produce continuous DVS input rather than the synchronous input produced by a computer monitor. The DC motor speed was controlled by PWM output from an Arduino Nano. We varied the speed of rotation under computer control and measured the output event rate $R$ using various $\tweakrefr$ settings.  Decreasing $\tweakrefr$ decreased the event rate. The 1\,ms snapshots show that increased refractory period erases repeated events and the trailing ON edge of the black dot. The summary plot of event rate versus refractory period shows that $\tweakrefr$ systematically affects the event rate and the effect is nearly linear. 

\subsubsection{Effect of bandwidth on signal and noise $\snr$}
\label{sec:snr_results}
\begin{figure}
    \centering
    \includegraphics[width=.7\columnwidth]{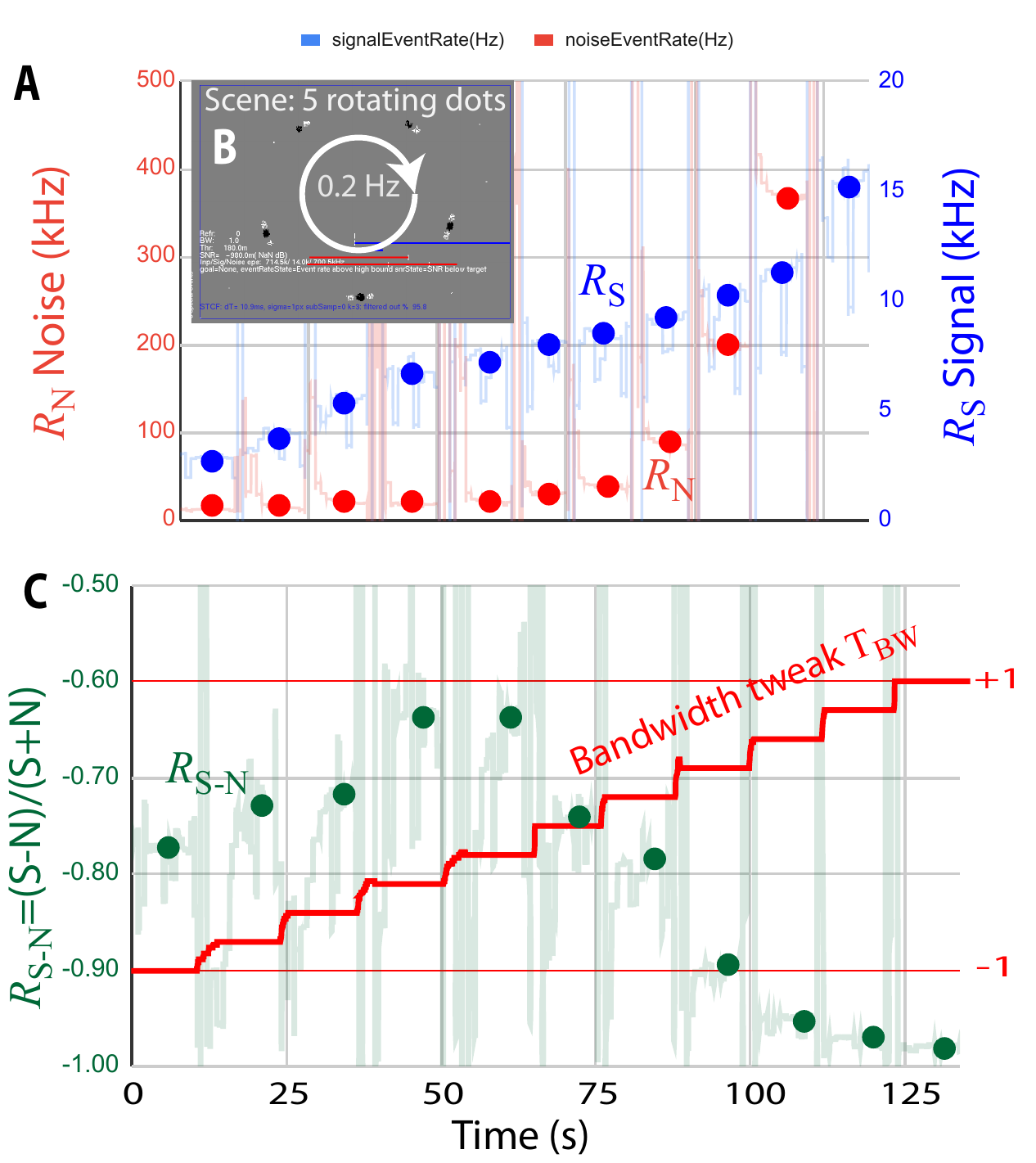}
    \caption{Measured signal and noise (both estimated from \eqref{eq:snr}) versus bandwidth tweak $\tweakbw$. Input is 5 rotating dots displayed on flicker-free monitor. Note large transients caused by tweak changes. }
    \label{fig:snr_vs_bandwidth}
\end{figure}

\vspace{-6pt}
Fig~\ref{fig:snr_vs_bandwidth} shows experimental results illustrating the concept of optimizing signal versus noise presented in Sec.~\ref{subsec:bwcontrol}. The input in this experiment was 5 rotating dots. As we increase the bandwidth tweak $\tweakbw$, the signal event rate $\rsignal$ increases but tends to saturate as it becomes high enough to capture the movements of the dots. The noise event rate $\rnoise$ continues to increase steeply. The signal rate appears to increase more steeply for the final $\tweakbw=1.0$ setting, but this increase is a byproduct of greatly increased noise, which is misclassified as signal. 
A maximum $\snr$ occurs for tweak values around zero. (The $\snr$ is still negative even at this peak, because for this scene, the signal event rate was always less than the noise event rate.) Big changes in event rate occur every time $\tweakbw$ is changed which makes the measurement quite difficult.

\subsection{Control experiments}
\label{sec:control_experiments}
This section presents results of control experiments.
\subsubsection{Bounding event rate by controlling threshold}
\label{sec:bound_event_rate_expt}
\begin{figure}
    \centering
    \includegraphics[width=.8\columnwidth]{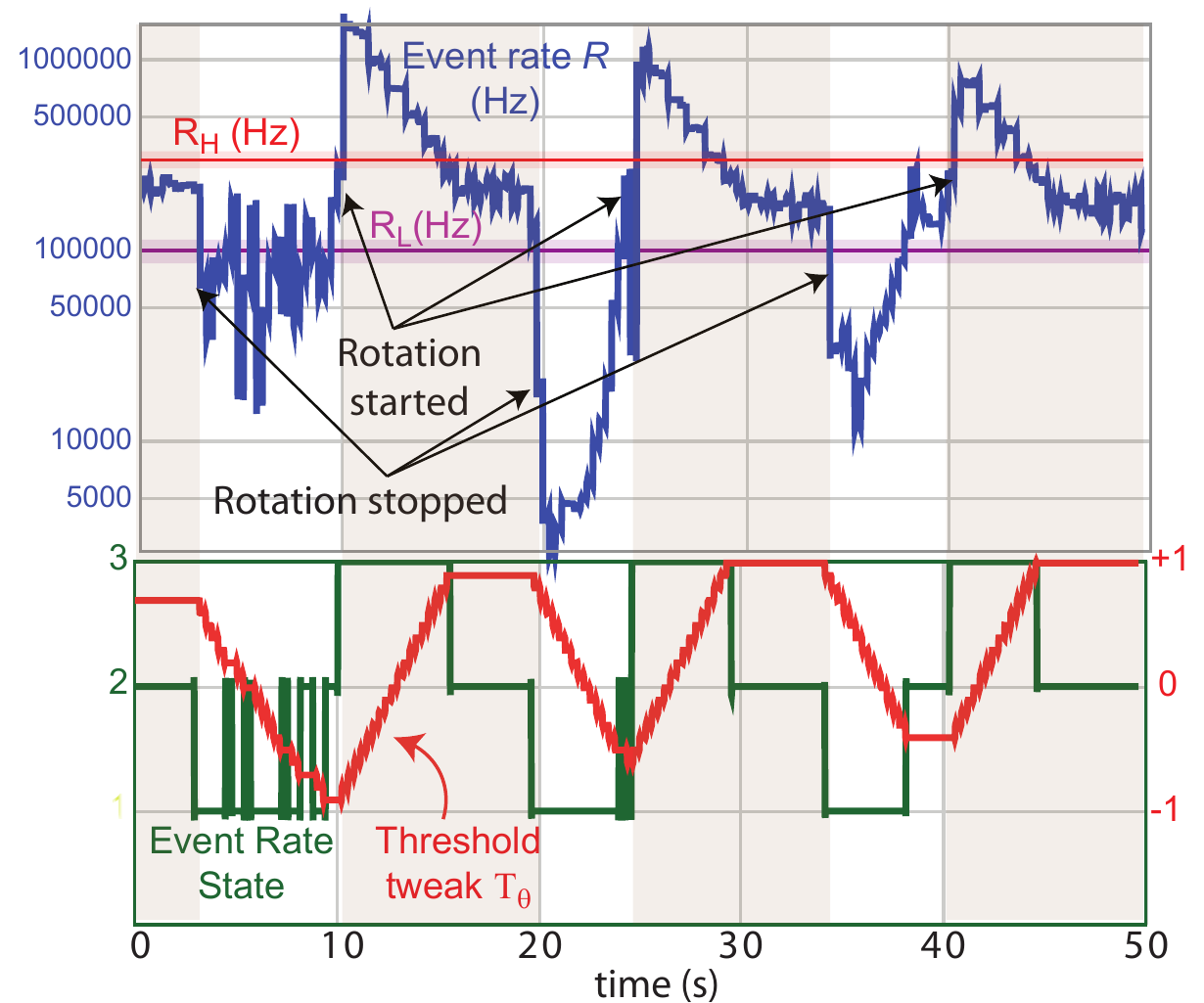}
    \caption{Fixed-step event rate bounding using threshold tweak $\tweakthr$ and $[\rlow,\rhigh]=[100,300]\text{ kHz}$. }
    \label{fig:event_rate_bang_bang_control_expt}
\end{figure}

\vspace{-6pt}
Fig.~\ref{fig:event_rate_bang_bang_control_expt} shows the result of an experiment on bounding event rate using fixed-step threshold control. The input was a single rotating dot for which we varied the speed of rotation (and occasionally stopped) to vary the event rate. We arbitrarily set $\rhigh=300\text{\,kHz}$ and $\rlow=100\text{\,kHz}$  for this experiment; in practice they would be determined by system-level considerations. We can see that the threshold tweak varies according to the control policy to bring the event rate within the $[\rlow,\rhigh]$ bounds.

\subsubsection{Limiting high event rate with refractory period}
\label{sec:refr_control_expt}
\begin{figure}
    \centering
    \includegraphics[width=.8\columnwidth]{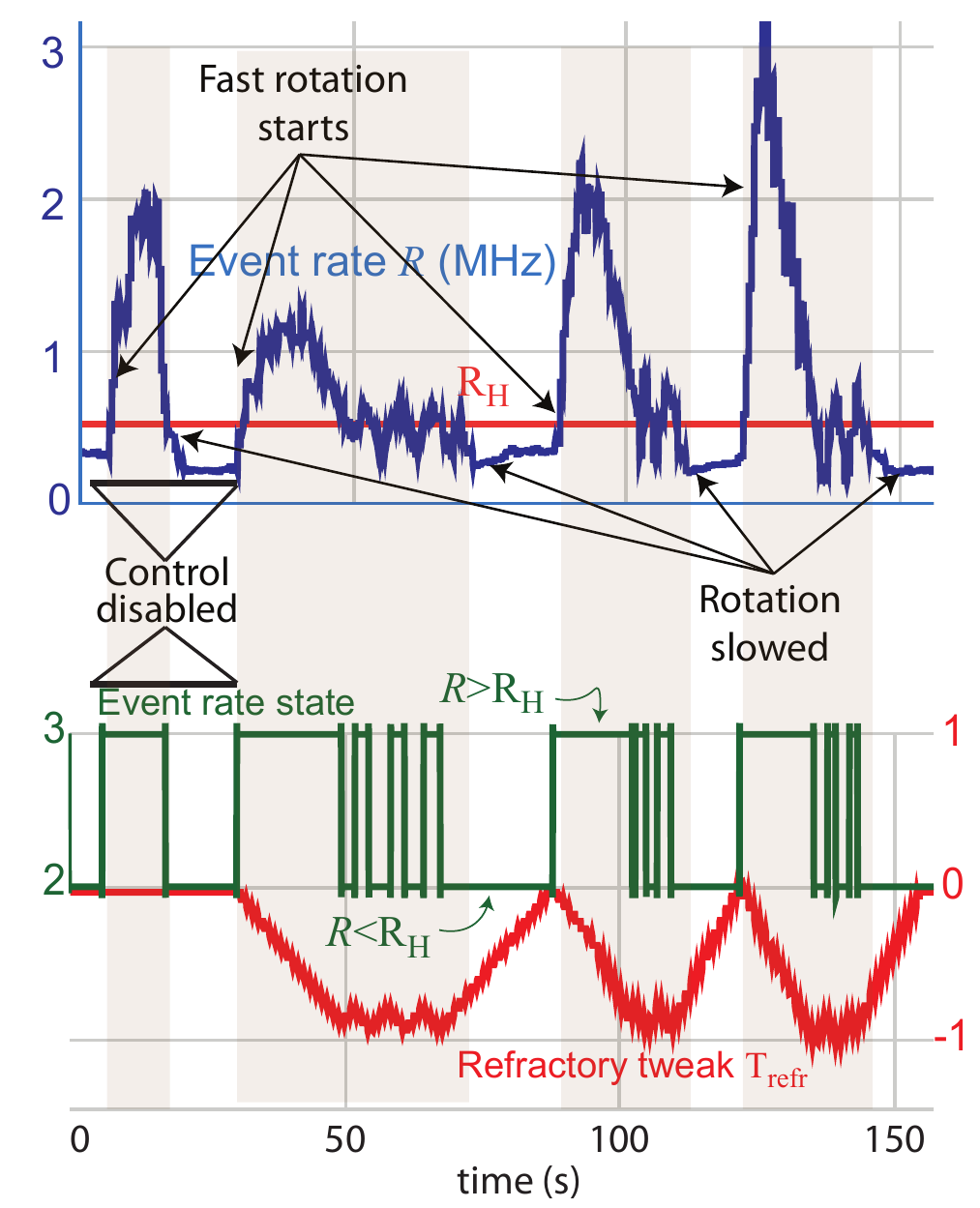}
    \caption{Fixed-step event rate limiting using refractory period tweak $\tweakrefr$ and $\rhigh=500\text{kHz}$.  }
    \label{fig:limiting_event_rate_expt}
\end{figure}

\vspace{-6pt}
Fig~\ref{fig:limiting_event_rate_expt} shows the results of an experiment to control event rate $R$ below a maximum limit $\rhigh$ by using $\tweakrefr$. We displayed input to the DVS where we could control the number of generated DVS events by varying the speed of the input. Initially, the control is disabled, and the $R$ goes up to over 2\,MHz. Then control is enabled. While $R\geq\rhigh$, $\tweakrefr$ is decreased. After 5s, $R$ is controlled to be under $\rhigh$. The cycle is repeated several times.  

\subsubsection{Regulating noise with bandwidth control}
\label{sec:limiting_noise_results}
\begin{figure}
    \centering
    \includegraphics[width=\columnwidth]{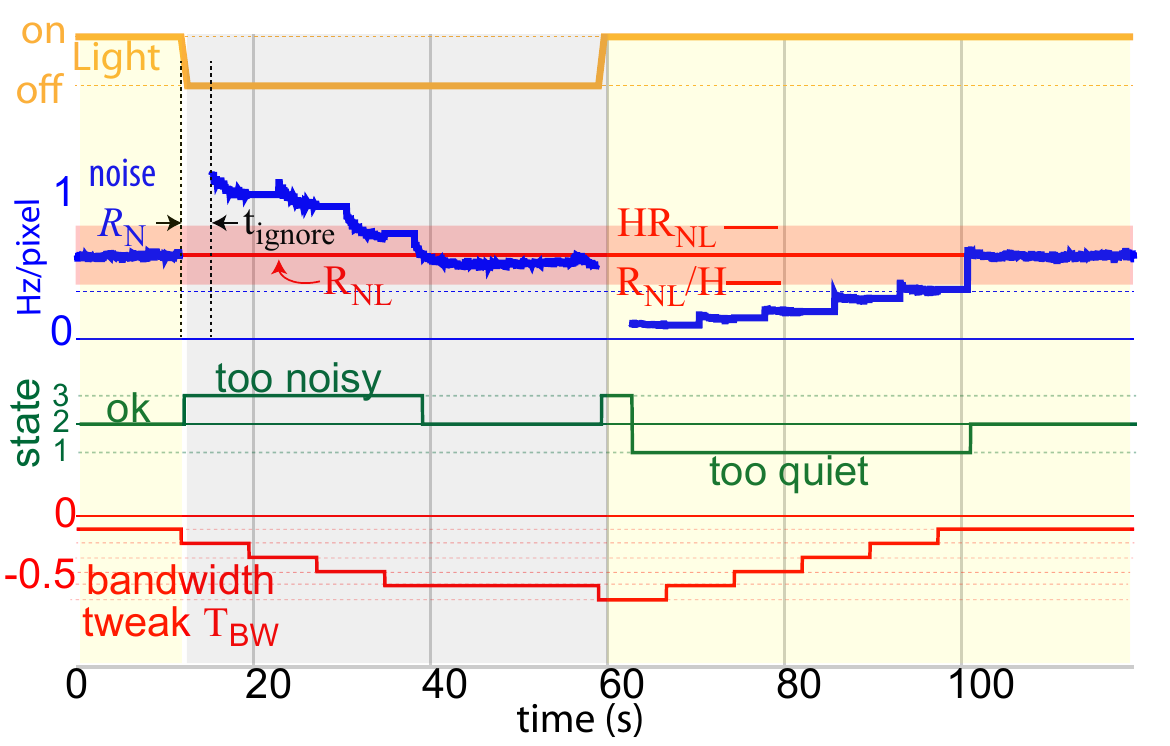}
    \caption{Noise regulation using bandwidth tweak $\tweakbw$.  }
    \label{fig:limiting_noise_expt}
\end{figure}

\vspace{-6pt}
Fig~\ref{fig:limiting_noise_expt} shows the results of an experiment to regulate noise per pixel $\rnoise$ close to $\rnoiselimit$ by using $\tweakbw$. 
In a stationary scene, we turned the light illuminating the scene off and then on again. Turning the light off increased the noise $\rnoise$, so the controller decreased $\tweakbw$ until $\rnoise$ went below $\rnoiselimit$ by the required hysteresis factor of $
\bbhys=1.5$. Then we turned the light back on. This caused $\tweakbw$ to increase back to its original value.
Thus the noise controller successfully regulates the DVS pixel shot noise.

\section{Discussion and conclusion}
\label{sec:conclusion}

\begin{table}[ht]
\scriptsize
\centering
\setlength{\tabcolsep}{0.5em}
    \begin{tabular}{L{1cm}|L{1.2cm}|L{1.5cm}|L{3.1cm}}
        \setlength{\parindent}{0pt}
        \textbf{Control} & \textbf{Bias} & \textbf{Advantages} & \textbf{Disadvantages} \\
        \hline
        Event rate & Threshold & Does not discard high frequency information. & Increased brightness quantization with higher threshold. Decreasing threshold causes event bursts.  \\
        \hline
       Event rate & Refractory period & Only minor control artifacts. & Increased refractory period discards high spatio-temporal frequency information. \\
        \hline
        Noise & Bandwidth & Controlled noise. & Large global event transient artifacts caused by $\tweakbw$ changes. Difficult to measure $\snr$ over short time scales. \\
        \hline
    \end{tabular}
    \caption{Comparing controllers.}
    \label{tab:comparison}
\end{table}

Table~\ref{tab:comparison} compares the control method advantages and disadvantages.
Regulating event rate is clearly practical. Either threshold or refractory period control can be used. Threshold control has the advantage of increasing sensitivity for low contrast features, but increased sensitivity increases noise. Additionally, changes of threshold (particularly decreases) introduce significant transient bursts of events. A more sophisticated controller based on a more precise model would allow sparser control action and so would limit the undesired transient effects. It would also speed up control, making it suitable for wider range of applications. A good first choice would be a proportional controller which we predict to perform well thanks to linear dependence between $\tweakthr$ and event rate. Proportional control may be particularly effective if one knows $R_0$ (from \eqref{eq:ratevsthreshold}) and are able to infer $\sens_\text{min}$ online. 

Changing refractory period does not cause transient control artifacts, but leads to loss of high spatio-temporal frequencies. For example, for our test case of a spinning black dot, a long refractory period erases the trailing ON edge of the dot. The effect of a long refractory period on textured objects would be to erase much of the detailed texture contained within the object, leaving only the leading edge and some of the internal structure.

Automatic control of signal and noise tradeoff is more difficult: 
Conceptually, a particular tradeoff of signal versus noise is arbitrary,
since the relative cost of admitting more noise or losing more high frequency signal depends on the application. 
Practically, it is hard to measure a metric like $\snr$ metric in a dynamically-changing scene. The measurements of Fig.~\ref{fig:snr_vs_bandwidth} were obtained with a well-controlled, unvarying stimulus where signal events were designed to be as constant as possible. In real scenarios, signals would usually change in an unpredictable way. Measuring $\rnoise$ to limit it is more practical, but each bandwidth tweak change introduces a lot of transient artifact events, because the global DC changes of photoreceptor output cause large bursts of noise events. Therefore, the controller must ignore the output until the transient disappears. Automatic noise regulation would be useful for long-term control, e.g., in surveillance and environmental monitoring, where the noise measurements would be reliable and bandwidth control actions would be infrequent, e.g. every few minutes.

The results in this paper on automatic feedback control of DVS camera pixel biases should be regarded as a starting point for future improvements.
The controllers are currently designed to work in isolation, but could be combined, \eg noise could be controlled by bandwidth and threshold. Bias control could incorporate time of day~\cite{Litzenberger2007-car-counting}, lighting, and temperature information. Absolute lighting information is available from sensors that include intensity output like DAVIS but could also incorporate a simple light sensor or a global photocurrent measurement such as in~\cite{dvs128}. Temperature is available from 
the camera's inertial measurement unit~\cite{Delbruck2014-imu}. These sensor readings can be fused with DVS output statistics for quicker and more robust bias control.

\subsection*{Acknowledgements}
This work was supported by Swiss SNF grant SCIDVS (200021\textunderscore185069). We thank reviewers for useful comments.

\renewcommand*{\bibfont}{\footnotesize}
\printbibliography

\end{document}